# Probabilistic Evaluation of Candidates and Symptom Clusterings for Multidisorder Diagnosis*

Thomas D. Wu
*MIT Laboratory for Computer Science*
*545 Technology Square, Cambridge, Massachusetts 02139*
`tdwu@lcs.mit.edu`

## Abstract

This paper derives a set of formulas for computing the probability of a symptom clustering, given a set of positive and negative findings. Symptom clusterings are produced by a recent method for multidisorder diagnosis that efficiently finds minimal candidates, or disorder combinations, to explain a given set of positive findings. Each symptom clustering represents a collection of such candidates. The probabilistic result for a symptom clustering derived in this paper therefore allows a large set of candidates to be probabilistically validated or pruned simultaneously.

The probability of a collection of candidates is then limited to obtain a special case: the probability of a single candidate. Unlike earlier results, the equation derived here allows the specification of positive, negative, and unknown symptoms and does not make assumptions about disorders not in the candidate.

## 1 Introduction

Diagnosis of multiple disorders is an important but computationally challenging problem. One approach to multidisorder diagnosis is set covering [1,6,7], which finds a candidate set of disorders that minimally explains a given set of symptoms. A recently developed method for computing set covers is the symptom clustering algorithm [8,9], which offers increased efficiency compared to existing candidate generation methods. Preliminary results on a large, real-world knowledge base (INTERNIST/QMR [3,4]) indicate that the symptom clustering algorithm yields near-exponential performance gains over candidate generation methods [9].

Nevertheless, the symptom clustering algorithm is categorical rather than probabilistic. This paper extends the symptom clustering framework to allow probabilistic information. This is important in fields such as medicine where the likelihood of diseases and causal influences ranges over several orders of magnitude. This paper derives a set of formulas for the probability of a symptom clustering. Since a symptom clustering may represent several candidates, the probabilistic quantity derived in this paper may be used to evaluate a large set of candidates simultaneously. This initial result, the probability of a symptom clustering, is then specialized to give a second result: the probability of a single candidate. This probability differs from earlier work in that it does not make assumptions about disorders not in the candidate or symptoms not in the given case.

---

*This research was supported in part by National Institutes of Health grant R01 LM04493 from the National Library of Medicine and by National Research Service Award T32 GM07753.

## 2  Background

In multidisorder diagnosis, we assume that the domain can be represented as sets of disorders and symptoms. The total set of disorders in the knowledge base is $D_K$, and the total set of symptoms is $S_K$. Each disorder can cause certain symptoms; this is the Effects-of relation. Conversely, each symptom can be caused by certain disorders; this is the Causes-of relation. A sample knowledge base, expressed in terms of these two inverse relations, might appear as follows:

| Disorder $d_i$ | Effects-of($d_i$) | Symptom $s_i$ | Causes-of($s_i$) |
|---|---|---|---|
| $d_1$ | $\{s_2, s_3, s_7\}$ | $s_1$ | $\{d_2, d_4, d_5\}$ |
| $d_2$ | $\{s_1, s_2, s_4, s_5, s_6\}$ | $s_2$ | $\{d_1, d_2, d_3\}$ |
| $d_3$ | $\{s_2, s_3, s_6\}$ | $s_3$ | $\{d_1, d_3, d_5\}$ |
| $d_4$ | $\{s_1, s_4, s_5\}$ | $s_4$ | $\{d_2, d_4, d_6\}$ |
| $d_5$ | $\{s_1, s_3, s_6\}$ | $s_5$ | $\{d_2, d_4\}$ |
| $d_6$ | $\{s_4, s_7\}$ | $s_6$ | $\{d_2, d_3, d_5\}$ |
|  |  | $s_7$ | $\{d_1, d_6\}$ |

The two major approaches to multidisorder diagnosis are belief networks [5] and set covering [1,6,7]. These two approaches pursue different goals. Belief networks compute probabilities of individual disorders $d$ while set covering methods compute plausible disorder combinations $D$. Thus, they differ in two respects: their emphasis on individual disorders versus disorder combinations, and their emphasis on probabilistic versus categorical results.

Belief networks determine probabilities using a distributed algorithm, with each disorder, symptom, or intermediate state performing a local computation and passing information via links to related disorders, symptoms, and states. In the belief updating process, a belief network computes the posterior probability of each disorder, given a set of positive findings $P$ and negative findings $N$. But these probabilities are for individual disorders; they do not indicate what combinations of disorders, or interpretations, are likely. An interpretation can be thought of as a truth assignment—either presence or absence—to each disorder in the knowledge base. Unfortunately, probabilities of individual disorders do not necessarily translate to plausible combinations of disorders in a straightforward way. For instance, if we merely assign "presence" to each disorder with a probability greater than 0.5 and "absence" to each disorder with a probability less than 0.5, the resulting interpretation may be highly unlikely. Or if we take the two or three most probable disorders, the result may also be highly unlikely.

Set covering methods, on the other hand, deal with entire combinations of disorders at a time. In this approach, these are called candidates instead of interpretations, and the goal is to find minimal candidates for a given set of symptoms. A set $C$ of disorders is a *candidate* for a set $P$ of positive findings if each symptom in $P$ can be explained by some disorder in $C$. A candidate is *minimal* if none of its subsets is a candidate for the given symptoms. For instance, if symptoms $s_1$, $s_2$, $s_3$, and $s_4$ were present, a minimal candidate would be $[d_1, d_2]$. This explains all four symptoms and is minimal because $d_1$ or $d_2$ alone would fail to explain them.

While it is true that belief networks have a limited way, called belief revision, to compute disorder combinations, this is limited to finding only the best and second-best interpretations. In order to obtain a more complete list of interpretations, we need to use the set covering approach. Unfortunately, set covering approaches suffer from two limitations: (1) they are combinatorially explosive, and (2) they are essentially categorical. The combinatorial problem has been addressed recently by developing a method called symptom clustering [8,9] to efficiently compute collections of candidates, instead of individual candidates. Each collection of candidates is represented by a single symptom clustering. Therefore, the search processes of generation, evaluation, and pruning can be conducted at the level of symptom clusterings, thereby reducing the combinatorics of the candidate search process. The second limitation, the need for probabilistic methods, is addressed in this paper. The probabilistic methods derived here may then be used in conjunction with the efficient symptom clustering representation to facilitate the set covering approach to multidisorder diagnosis.

## 3 Symptom Clusterings

We now introduce relevant terminology for the symptom clustering approach. The actual construction of a symptom clustering is beyond the scope of this paper and is reported elsewhere [9]. In this paper, we are concerned only with evaluating the likelihood of a symptom clustering after it has been constructed.

In symptom clustering, we begin with a set $P$ of positive findings and a set $N$ of negative findings. Any symptom not in $P$ or $N$ is assumed to be unknown; it could be either present or absent. The positive findings are partitioned into *clusters*. All symptoms in a cluster are hypothesized to be caused by the same disorder, and the set of disorders that can explain a cluster is called its *differential diagnosis*. A cluster $S$ and differential diagnosis $D$ together form a *problem area* or *task*, symbolized as $(S \leftarrow D)$. This means the situation where every symptom in $S$ is caused by some disorder in differential diagnosis $D$. A *clustering* $C$ is composed of a set of tasks: $(S_1 \leftarrow D_1), \ldots, (S_n \leftarrow D_n)$.

Below, we show an example of a symptom clustering for $P = \{s_1, s_2, s_3, s_4\}$. This clustering follows from the knowledge base presented above.

|  | Cluster $S$ | Differential $D$ |
|---|---|---|
| Problem area 1: | $s_1$ and $s_4$ | $d_2$ or $d_4$ |
| Problem area 2: | $s_2$ and $s_3$ | $d_1$ or $d_3$ |

This clustering indicates a situation where two problem areas are present, one accounting for symptoms $s_1$ and $s_4$, and one for symptoms $s_2$ and $s_3$. The first cluster can be explained by disorder $d_2$ or $d_4$; the second by disorder $d_1$ or $d_3$. This is because $d_2$ and $d_4$ are in the Causes-of sets of both $s_1$ and $s_4$; likewise, $d_1$ and $d_3$ are causes for both $s_2$ and $s_3$.

A symptom clustering is efficient because it represents a collection of candidates rather than a single candidate. The candidates entailed by a symptom clustering are the cartesian product of the differential diagnoses:

$$\text{Cands}(\mathcal{C}) = \bigtimes_{i=1}^{n} D_i$$

Thus, in our example, a single clustering entails four candidates, namely, $[d_2, d_1]$, $[d_2, d_3]$, $[d_4, d_1]$, and $[d_4, d_3]$. Of course, as the knowledge base grows, the efficiency gains can be much greater. By the way in which differential diagnoses are constructed (for details, see [9]), we may assume that no two differential diagnoses share the same disorder. This requirement helps ensure that the candidates entailed by a clustering are minimal.

The question we pose in this paper is this: What is the probability that one of the candidates entailed by clustering $\mathcal{C}$ is present, given positive findings $P$ and negative findings $N$? We express this value as $p(\text{Cands}(\mathcal{C}) \mid P^+ N^-)$, where the "+" superscript indicates that all elements of the underlying set are present, and the "−" superscript indicates that all elements of the underlying set are absent. Note that this is the probability of a disjunction: that one or more candidates in $\text{Cands}(\mathcal{C})$ is present. In turn, a candidate is present if all of its component disorders are present; this is a conjunction. Technically, then, we should write our desired goal as $p\left(\bigvee_{C \in \text{Cands}(\mathcal{C})} C^+ \mid P^+ N^-\right)$. However, for simplicity, we shall use the notation $p(\text{Cands}(\mathcal{C}) \mid P^+ N^-)$ instead. Note further that if a candidate is present, this does not preclude other disorders from also being present. Thus, if disorders $d_1$, $d_2$, and $d_3$ are present, then candidate $[d_1, d_2]$ is considered to be present. Equivalently, this means that more than one candidate in $\text{Cands}(\mathcal{C})$ can be present at the same time.

## 4 Probability of a Symptom Clustering

In this section, we derive a set of formulas for the probability of a symptom clustering $p(\text{Cands}(\mathcal{C}) \mid P^+ N^-)$. These formulas rely upon various probabilities given by the knowledge base. We assume the knowledge base has a prior probability for each disorder $d$, so that $p(d^+)$ is the prior probability of $d$ being present, and $p(d^-) = 1 - p(d^+)$ is the prior probability of $d$ being absent. We also assume that the knowledge base has causal probabilities between all pairs of disorders and symptoms. The causal probability between $d$ and $s$ is $c_{ds} = p(s \mid \text{only } d)$, that is, the probability that $s$ would occur if $d$ were present and all other disorders in the knowledge base were absent. If $s$ is not a possible effect of $d$ in the knowledge base, then $c_{ds} = 0$.

With these probabilities, we can now derive the probability of a symptom clustering given positive findings $P$ and negative findings $N$. This probability can be computed using the definition of conditional probability:

$$p(\text{Cands}(\mathcal{C}) \mid P^+ N^-) = \frac{p(P^+ N^- \text{Cands}(\mathcal{C}))}{p(P^+ N^-)} \qquad (1)$$

The right-hand side of this equation consists of a numerator and denominator. We consider these factors separately in the next two subsections.

## 4.1 Numerator

Computing the probability of the numerator $p(P^+N^-\text{Cands}(\mathcal{C}))$ requires two techniques. First, we use an inclusion-exclusion strategy. This strategy turns positive findings into an alternating sum of terms, each of which assumes some of these findings are actually absent:

$$p(P^+N^-\text{Cands}(\mathcal{C})) = \sum_{S \in 2^P} (-1)^{|S|} p((S \cup N)^-\text{Cands}(\mathcal{C})) \qquad (2)$$

In this equation, $2^P$ denotes the power set of $P$, and $(S \cup N)^-$ denotes the event that all symptoms in $(S \cup N)$ are absent. From now on, we use the symbol $N' = (S \cup N)$ to denote this augmented set of absent findings.

The second technique is to factor the term in the above summation into $(n+1)$ parts: $n$ of which correspond to a cluster in $\mathcal{C}$, and one of which corresponds to disorders not in any differential of $\mathcal{C}$. Let us call the $n$ differentials $D_i$ and call the remaining diseases $D^* = D_K - \bigcup_{i=1}^n D_i$, where all $D_K$ is the set of all disorders in the knowledge base. This factorization yields

$$p(N'^-\text{Cands}(\mathcal{C})) = \left[\prod_{i=1}^n p\left(\bigwedge_{d \in D_i}(N' \not\leftarrow d) \wedge \bigvee_{d \in D_i} d^+\right)\right] p\left(\bigwedge_{d \in D^*}(N' \not\leftarrow d)\right) \qquad (2a)$$

Here, the notation $(N' \not\leftarrow d)$ means that individual disorder $d$ does not cause any symptoms in $N'$. The notation $\bigwedge_{d \in D_i}(N' \not\leftarrow d)$ means that all disorders in the differential diagnosis $D_i$ fail to cause any symptoms in $N'$. And the notation $\bigvee_{d \in D_i} d^+$ means that some disorder $d$ in the differential diagnosis $D_i$ is present.

Consider one of the $n$ differentials. A disorder $d$ in $D_i$ will fail to cause any symptom in $N'$ if (1) it is absent, or (2) if it is present but still fails to cause any of the symptoms. However, we must disallow one instance where $D_i$ fails to cause $N'$, namely, when all disorders in $D_i$ are absent. This situation is excluded by the fact that $\bigvee_{d \in D_i} d^+$ must be true, so at least one disorder in $D_i$ is present. Thus, we subtract the probability that all disorders in $D_i$ are absent. Altogether, this yields the following result:

$$p\left(\bigwedge_{d \in D_i}(N' \not\leftarrow d) \wedge \bigvee_{d \in D_i} d^+\right) = \prod_{d \in D_i}\left(p(d^+) \prod_{s \in N'}(1 - c_{ds}) + p(d^-)\right) - \prod_{d \in D_i} p(d^-) \qquad (2b)$$

Now consider the set $D^*$ of remaining disorders, those not in any differential of $\mathcal{C}$. The reasoning here is the same as before, except that we do not require that some disorder in $D^*$ be present:

$$p\left(\bigwedge_{d \in D^*}(N' \not\leftarrow d)\right) = \prod_{d \in D^*}\left(p(d^+) \prod_{s \in N'}(1 - c_{ds}) + p(d^-)\right) \qquad (2c)$$

In summary, to obtain the numerator of (1), substitute equations (2b) and (2c) in (2a), and in turn substitute this in equation (2).

465## 4.2 Denominator

The denominator expresses the marginal probability of a set of positive and negative findings. As with the first term in the numerator, this may be computed by an inclusion-exclusion principle that considers an alternating sum of terms with only negative findings [2]:

$$p(P^+ N^-) = \sum_{S \in 2^P} (-1)^{|S|} \prod_{d \in D_K} \left( p(d^+) \prod_{s \in S \cup N} (1 - c_{ds}) + p(d^-) \right) \qquad (3)$$

where $2^P$ is the power set of $P$, and a $D_K$ is all disorders in the knowledge base.

## 4.3 Computational Complexity

The computational complexity of the above set of formulas is shown in the table below:

| | |
|---|---|
| Numerator term 1: | $\mathcal{O}(2^\mathcal{P} \mathcal{N} \mathcal{D})$ |
| Numerator term 2: | $\mathcal{O}(\mathcal{N} \mathcal{D})$ |
| Denominator: | $\mathcal{O}(2^\mathcal{P} (\mathcal{P} + \mathcal{N}) \mathcal{D})$ |

where $\mathcal{P}$ is the number of positive findings, $\mathcal{N}$ is the number of negative findings, and $\mathcal{D}$ is the number of disorders in the knowledge base. The denominator term dominates, so the overall computational complexity is $\mathcal{O}(2^\mathcal{P}(\mathcal{P} + \mathcal{N})\mathcal{D})$, which is exponential in the number of positive findings but linear in the number of negative findings and size of the knowledge base. Since the number of positive findings is often relatively small for a particular case, this complexity may be manageable for several real-world knowledge bases. Finally, note that the numerator has roughly the same complexity as the denominator, so there seems to be little advantage to computing relative probabilities between clusterings as opposed to absolute probabilities.

## 5 Probability of a Single Candidate

The probability given in the previous section evaluates a *collection* of candidates, namely the cartesian product of the differential diagnoses in a symptom clustering. In this section, we specialize this formula to obtain the probability of a *single* candidate $C$. We merely assume that the differential diagnoses $D_i$ in some symptom clustering are all singletons. Then, the candidate $C$ is the only element in the cartesian product:

$$\bigtimes_{i=1}^{n} D_i = \{C\}$$

The result for this limiting case is:

$$p(C^+ \mid P^+ N^-) =$$

$$\sum_{S \in 2^P} \frac{(-1)^{|S|} \prod_{d \in C} \left( p(d^+) \prod_{s \in S \cup N} (1 - c_{ds}) \right) \prod_{d \in D_K - C} \left( p(d^+) \prod_{s \in S \cup N} (1 - c_{ds}) + p(d^-) \right)}{p(P^+ N^-)} \qquad (4)$$

The formula for the denominator remains unaltered and can be found in equation (3). Note that in the absence of any evidence, $P = N = \emptyset$, this expression reduces to the expected result for the prior probability of a candidate:

$$p(C^+) = \prod_{d \in C} p(d^+)$$

Previous work in set covering has also considered the probability of a single candidate, such as work by Peng and Reggia [6]. However, their result simplifies the computation by introducing assumptions about disorders not in the candidate and symptoms not in the set of positive findings. For instance, they assume that all disorders not in the candidate are absent and that all symptoms not in the set of positive findings are also absent. This is equivalent to finding the probability for the event "only $D$ present" given "only $P$ present". Their result is reproduced here:

$$p(\text{only } C^+ \mid \text{only } P^+) = \frac{\prod_{s \in P}\left(1 - \prod_{d \in C}(1 - c_{ds})\right) \prod_{s \in S_K - P}\left(\prod_{d \in C}(1 - c_{ds})\right)}{\prod_{d \in D_K - C} p(d^-)}$$

Thus, there are essentially two assumptions required to obtain Peng and Reggia's result: (1) only the disorders in $C$ are present and (2) only the symptoms in $P$ are present. These assumptions may not valid in some domains. For example, in medicine, people often have minor illnesses, such as the common cold. But the first assumption requires that the patient not have any diseases, however minor, except those that are included in the candidate $C$. The second assumption is even less reasonable. It requires that we know the presence or absence of every finding in the knowledge base. This is an infrequent situation in most domains. For example, this assumption forces us to decide, even before a chest X-ray has been performed, whether the result is going to be positive or negative. In most domains, especially those requiring reasoning about uncertainty, the status of most evidence is unknown.

In this paper, on the other hand, both of these assumptions are relaxed. To relax the first assumption, we have defined a candidate to be present whenever all of its component disorders are present, regardless of whether other disorders are present or absent; the status of these other disorders is simply unknown. To relax the second assumption, we allow two sets of findings, positive and negative, to be specified. Findings in the knowledge base not in either category are assumed to be unknown. This differs from the Peng and Reggia approach where only one set of findings, the present ones, are specified, while all other findings are assumed to be absent.

## 6 Conclusion and Further Work

This paper has derived equations for the probability of the collection of candidates entailed by a symptom clustering and for the probability of a single candidate, given a set of positive and negative findings. The computational advantage of the first result is that a symptom clustering may entail several candidates, so that the feasibility of a large set of candidates can be determined simultaneously. The second result relaxes assumptions made by other researchers about disorders not in the candidate or symptoms not in the given case.

The complexity of these equations is exponential in the number of positive findings but linear in the number of negative findings and in the size of the knowledge base. Although the number of positive findings is often limited, this exponential complexity may still present problems in some cases. Further research in this area might approximate the probability of a symptom clustering or determine an upper bound for these probabilities. For many purposes, such as guiding search, approximate values and upper bounds would be sufficient. A probabilistic symptom clustering algorithm might then increase the performance gains already obtained with the categorical algorithm.

# References

[1] Johan de Kleer and Brian C. Williams. Diagnosing multiple faults. *Artificial Intelligence*, 32:97–130, 1987.

[2] David Heckerman. A tractable inference algorithm for diagnosing multiple diseases. In *Fifth Workshop on Uncertainty in Artificial Intelligence*, pages 174–181, 1989.

[3] R. A. Miller, M. A. McNeil, S. M. Challinor, F. E. Masari, Jr., and J. D. Myers. The Internist-1/Quick Medical Reference project—status report. *Western Journal of Medicine*, 145:816–822, 1986.

[4] Randolph A. Miller, Harry E. Pople, Jr., and Jack D. Myers. Internist-1, An experimental computer-based diagnostic consultant for general internal medicine. *New England Journal of Medicine*, 307:468–476, 1982.

[5] Judea Pearl. *Probabilistic Reasoning in Intelligent Systems: Networks of Plausible Inference*. Morgan Kaufman, 1988.

[6] Yun Peng and James A. Reggia. A probabilistic causal model for diagnostic problem solving—Part I: Integrating symbolic causal inference with numeric probabilistic inference. *IEEE Transactions on Systems, Man, and Cybernetics*, SMC-17:146–162, 1987.

[7] Raymond Reiter. A theory of diagnosis from first principles. *Artificial Intelligence*, 32:57–96, 1987.

[8] Thomas D. Wu. Symptom clustering and syndromic knowledge in diagnostic problem solving. In *Proceedings, Thirteenth Symposium on Computer Applications in Medical Care*, pages 45–49, 1989.

[9] Thomas D. Wu. Efficient diagnosis of multiple disorders based on a symptom clustering approach. In *Proccedings, Eighth National Conference on Artificial Intelligence*, 1990 (to appear).